\RequirePackage{amsmath}
\documentclass[runningheads]{llncs}
\usepackage[T1]{fontenc}
\usepackage{graphicx}

\usepackage{cite}
\usepackage{amssymb,amsfonts}
\usepackage{algorithmic}
\usepackage{textcomp}
\usepackage{xcolor}
\usepackage{algorithm}
\usepackage{algorithmic}
\usepackage{booktabs}
\usepackage{textcomp, soul, xspace}
\usepackage{tikz}
\usepackage{comment}
\usepackage{amsmath, amssymb}
\usepackage{mfirstuc}
\usepackage{multirow}
\usepackage{times}  % DO NOT CHANGE THIS
\usepackage{helvet}  % DO NOT CHANGE THIS
\usepackage{courier}  % DO NOT CHANGE THIS
\usepackage[hyphens]{url}  % DO NOT CHANGE THIS
\usepackage{graphicx} % DO NOT CHANGE THIS
\usepackage{makecell}
\usepackage{censor}
\usepackage{balance}

\newcommand{\etal}{{\em et al.}\xspace}
\newcommand{\eg}{{\em e.g.,}\xspace}
\newcommand{\ie}{{\em i.e.,}\xspace}

\newcommand{\BfPara}[1]{\noindent\textbf{#1.}\xspace}

\begin{document}

\title{Identifying Cyberbullying Roles in Social Media}

\author{Manuel Sandoval\inst{1}\orcidID{0009-0009-8590-8811},\\
Mohammed Abuhamad \inst{1}\orcidID{0000-0002-3368-6024},\\
Patrick Furman \inst{1}\orcidID{0009-0009-9328-1481},\\
Mujtaba Nazari \inst{1}\orcidID{0009-0003-4731-1789},\\
Deborah L. Hall \inst{2}\orcidID{0000-0003-2450-3596}, \\\and
Yasin N. Silva \inst{1}\orcidID{0000-0003-1852-1683}
}

\authorrunning{M. Sandoval et al.}

\institute{Loyola Chicago University, Chicago IL 60660, USA\\
\email{\{msandovalmadrigal,mabuhamad,pfurman,mnazari,ysilva1\}@luc.edu}\\
\and
Arizona State University, Glendale AZ 85306, USA\\
\email{d.hall@asu.edu}
}

\maketitle              

\begin{abstract}
Social media has revolutionized communication, allowing people worldwide to connect and interact instantly. However, it has also led to increases in cyberbullying, which poses a significant threat to children and adolescents globally, affecting their mental health and well-being. It is critical to accurately detect the roles of individuals involved in cyberbullying incidents to effectively address the issue on a large scale. 
This study explores the use of machine learning models to detect the roles involved in cyberbullying interactions. After examining the AMiCA dataset and addressing class imbalance issues, we evaluate the performance of various models built with four underlying LLMs (\ie BERT, RoBERTa, T5, and GPT-2) for role detection. 
Our analysis shows that oversampling techniques help improve model performance. 
The best model, a fine-tuned RoBERTa using oversampled data, achieved an overall F1 score of $83.5$\%, increasing to $89.3$\% after applying a prediction threshold. The top-2 F1 score without thresholding was $95.7$\%. Our method outperforms previously proposed models.
After investigating the per-class model performance and confidence scores, we show that the models perform well in classes with more samples and less contextual confusion (\eg Bystander Other), but struggle with classes with fewer samples (\eg Bystander Assistant) and more contextual ambiguity (\eg Harasser and Victim). 
This work highlights current strengths and limitations in the development of accurate models with limited data and complex scenarios. 

\keywords{cyberbullying \and role detection \and social media \and LLM.}
\end{abstract}

\section{Introduction}
With the rise of social media, cyberbullying has become a widespread issue that affects young people worldwide.
Bullying, both conventional and online, has been the focus of numerous studies in the social sciences. Cyberbullying takes a variety of forms, such as spreading rumors, negative statements about race, gender, physical appearance, disability, or religion, humiliation, and threats of violence in public posts and comments. Cyberbullying is a serious issue with known harmful consequences, including psychological and social problems~\cite{teng2023cyberbullying}. 
Whereas many studies have focused on detecting cyberbullying incidents (\eg content), relatively fewer have been conducted to identify the roles of individuals/users involved in cyberbullying interactions \cite{rathnayake-etal-2020-enhancing, jacobs2022automatic}. Previous research has considered the following main roles: victim, bully, bystander assistant, bystander defender, and outsider. 
There are several critical benefits of accurate cyberbullying role detection: (1) it enables psychology researchers to better understand the dynamics of cyberbullying, \eg studying the underlying motivations and behaviors of bullies and temporal patterns of bystander-defender activity, (2) it can provide key information to help social media platforms implement targeted measures, \eg providing support and counseling resources for victims, (3) it can help social media platforms implement educational initiatives that raise awareness among bystanders about their role in enabling/resolving cyberbullying incidents and the importance of reporting, and (4) a better understanding of role dynamics enables the development of effective detection models for cyberbullying and anti-bullying.

To enable accurate role identification on a large scale, machine learning techniques could help identify patterns and indicators of cyberbullying behavior. The challenges of implementing such techniques include the need for large amounts of labeled data and addressing many scenarios where the roles of cyberbullying can overlap or the context is limited.\emph{This study addresses the task of identifying cyberbullying roles in social media interactions and sheds light on the merits and limitations of existing methods and datasets, paving the way for future research in this area.}
To this end, we examine and process the AMiCA dataset~\cite{van2018automatic} and employ oversampling methods to address the challenge of the imbalanced nature of the dataset.
We then develop and evaluate the performance of various machine learning models that are based on four large language models (LLMs): BERT, RoBERTa, T5, and GPT-2. Moreover, we compare the models we implemented with previously proposed role-detection approaches.

Our results indicate that providing context and employing oversampling significantly enhance the performance of models. 
Among other models, the fine-tuned RoBERTa model trained on oversampled data achieves an F1 score of $83.5$\% and a top-2 F1 score of $95.7$\%. The top-2 result indicates the probability of having the correct class in the top two predictions. Achieving this high top-2 F1 score prompted the investigation of the models' performance on a per-class granularity and the analysis of common cases of wrong prediction. 
For example, using a $25$-th percentile confidence score of the victim samples as a threshold for valid predictions increased the F1-score of fine-tuned RoBERTa to $89.3$\% (\ie $89.3-83.5 =5.8$\% improvement), with a rejection rate of $16.4$\% for comments. 

Our analysis also emphasizes the importance of training data and the embedded context within samples when building models to detect cyberbullying roles. The implemented models tend to exhibit strong performance when there are a large number of samples for a particular class, but encounter difficulties when there are fewer samples available. 

\BfPara{Contributions} The contributions of this study are twofold:
\begin{enumerate}
    \item Implementing and evaluating different strategies to build machine learning models for identifying the roles of individuals involved in cyberbullying incidents. These strategies involve processing the data, handling the class imbalance, training various models, and evaluating and analyzing their performance.

    \item Providing insights into the challenging nature of the cyberbullying role identification task and the limitations of the AMiCa dataset in addressing role overlap and extended conversational context.

\end{enumerate}

\section{Related Work} \label{sec:related_work}
Cyberbullying, in its many forms, has received considerable empirical attention within the social sciences, with much of the focus on the detrimental impact of cyberbullying on psychological and social outcomes of those involved~\cite{teng2023cyberbullying}. To protect internet users, particularly youth and adolescent users, from significant negative mental and psychosocial consequences, researchers have begun to develop frameworks for understanding and identifying the roles of different users who engage in or witness cyberbullying.

\BfPara{Cyberbullying Roles in the Social Sciences}
Research on both traditional bullying and cyberbullying has identified several distinct roles, including \textit{victim}, \textit{bully}, \textit{bully assistant}, \textit{defender of the victim}, and \textit{bystander}~\cite{salmivalli_bullying_1996}, each of which carries out a specific behavior that can influence the cyberbullying interaction. 
For instance, bystanders can reinforce the bully's actions, given that inaction can convey explicit or implicit cues that bullying is acceptable, funny, or even entertaining \cite{salmivalli_participant_2014}. Crucially, in an online setting, the cyberbullying-bystander feedback loop can manifest in actions specific to the platform, such as by providing reinforcing comments or by utilizing platform-specific features (\eg likes (Facebook, Instagram), upvotes (Reddit), or re-blogging (X, formerly known as Twitter). Garnering more followers can also function as a behavior-affirming signal for the cyberbully. Indeed, in previous research, cyber-bystanders who encouraged cyberbullying by reinforcing or assisting the aggressors ranked higher in the justification of violence than cyber-bystanders who defended or supported the victim \cite{Orue2023}. Additionally, users who helped reinforce a cyberbully had the highest scores in a measure of cyberbullying perpetration, indicating that those who support the aggressors are also likely to be the perpetrators in other cyberbullying interactions \cite{Orue2023}. Researchers have proposed that the choice to reinforce the cyberbully or support the victim is determined by a mix of personal and societal norms \cite{Dang_2022}. In terms of prevention, cyber-bystanders, \ie users who witness cyberbullying interactions, can play an active and key role in potential intervention. That is, bystanders have the capacity to intervene to support the victim and help alleviate the negative effects of bullying \cite{bastiaensens2015can}.

\BfPara{Cyberbullying Detection via Machine Learning}
Many studies have been proposed that apply off-the-shelf solutions, \eg SVM, Naïve Bayes, and Logistic Regression, to binary classification (bullying versus non-bullying) \cite{MDavar2012, xu_2012}. Dadvar and Eckert studied four deep learning architectures, CNN, LSTM, BiLSTM, and BiLSTM with attention on a cyberbullying-labeled YouTube dataset that included 54k posts and 4k users \cite{dadvar2018cyberbullying}. Cheng \etal included network-related content such as user profile information, likes, and follows to identify cyberbullying \cite{cheng_pi-bully_2019}. In both studies by Cheng \etal \cite{cheng_hierarchical_2019} and \cite{lu_cheng_2021_hant}, they modeled temporal dynamics using a hierarchical representation of social media sessions, where a session is composed of a sequence of comments, and a comment is a sequence of words. Other researchers have integrated network-related content, video, images, and time-related components into all-in-one deep learning architectures \cite{singh_toward_2017, wang_multi-modal_2020}. Ziems \etal \cite{ziems_agg_rep_int_visb_2020} collected a new dataset for cyberbullying detection based on X (formerly Twitter) that attempts to apply the definition of offline bullying, \ie the bullying interactions should contain aggressive language, be repetitive, contain harmful intent, be visible to peers, and present a power imbalance between the attacker and target.

\BfPara{Cyberbullying Role Detection via Machine Learning}
The area of cyberbullying role detection is relatively unexplored. To the best of our knowledge, the only models that address this problem are the ones by Jacobs \etal \cite{jacobs2022automatic} and Rathnayake \etal \cite{rathnayake-etal-2020-enhancing}, each of which is included in the present performance evaluation. Rathnayak \etal  \cite{rathnayake-etal-2020-enhancing} used the AMiCA dataset~\cite{van2018automatic} to develop a DistilBERT-based ensemble model \cite{sanh2020distilbert} to classify cyberbullying roles. While the authors report that their algorithm (\textit{OffensEval}) achieves an F1 score of $83$\%, their evaluation only considered 4 of the 5 roles in the AMiCA dataset. We found that OffensEval’s performance decreases significantly when considering all of the roles. 
Jacobs \etal \cite{jacobs2022automatic} used the AMiCA data to investigate multiple algorithmic configurations including single-algorithm classifiers (reporting $55$\% as the best F1 result with English data and Logistic Regression and $54$\% as the best score with Dutch data and Logistic Regression, SVM was the second best performing classifier), ensemble classifiers (reporting $55$\% as the best F1 result with English data using the Cascading approach), and transformer-based pretrained language models (reporting $55$\% as the best F1 score using RobBERT with Dutch data and $60$\% as the best score using RoBERTa with English data). The AMiCA dataset (Question-Answer pairs from AskFM) is the only labeled social media dataset that includes cyberbullying role labels. This dataset considers 5 cyberbullying roles: Harasser, Victim, Bystander Defender, Bystander Assistant, and Bystander Other. More recently, Hamlett \etal \cite{Hamlett_Powell_Silva_Hall_2022} proposed an Instagram dataset that includes a wide array of labels (including cyberbullying roles). This dataset, however, only contains 100 social media sessions making it difficult to integrate in robust ML model development.

\begin{figure*}[t]
\centering
\includegraphics[width=1.0\textwidth]{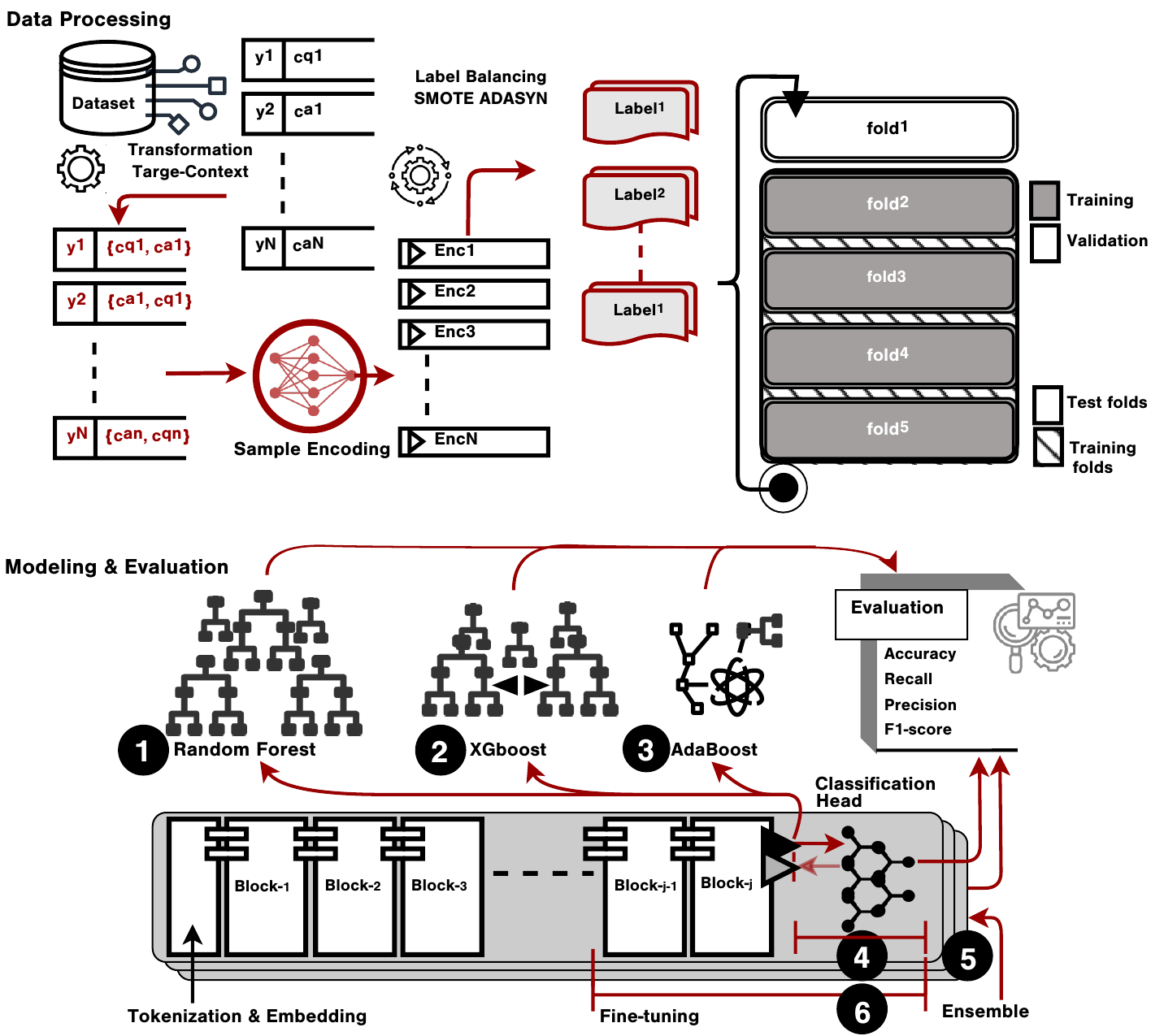}
\caption{Modeling Pipeline: Dataset is processed, and samples are transformed to target/context form and then processed to handle class imbalance. Using 10-fold cross-validation, LLMs are employed/evaluated using various methods for role detection. }
\label{fig:imp_path}
\end{figure*}

\section{Methods} \label{sec:methods}
Figure~\ref{fig:imp_path} provides an outline of our framework. This section describes the methods and design considerations.

\BfPara{Problem Definition} 
Let $C \in \{p_1, p_2, p_3, \dots, p_n \}$ be a corpus of $n$ samples, where a given $i\text{-th}$ sample $p_i = \{c_i, t_i\}$ is a pair of context $c_i$ and target $t_i$ text comments (corresponding to a Q\&A pair  from the dataset). 
Let $c_i = \{w_1^{c_i}, w_2^{c_i}, w_3^{c_i}, \dots, w_l^{c_i}\}$ and $t_i = \{w_1^{t_i}, w_2^{t_i}, w_3^{t_i}, \dots, w_l^{t_i}\}$ be the token representations of $c_i$ and $t_i$, respectively, for a length of $l$ tokens.
Let $Y \in \{y_1, y_2, y_3, \dots, y_n \}$ be the associated labels for samples in $C$, where a given
$i\text{-th}$ sample $p_i\rightarrow y_i$ such that $y_i = \{y_i^{c}, y_i^{t}\}$ (\ie $y_i^{c}$ and $y_i^{t}$ correspond to the labels associated with $c_i$ and $t_i$, respectively). For any given label $y_i^{x} \in \{0, 1, 2, 3, 4\}$ with values of $0$, $1$, $2$, $3$, and $4$ representing the labels \textit{harasser}, \textit{victim}, \textit{bystander-defender}, \textit{bystander-assistant}, and \textit{bystander-other}, respectively.

Let $f(p) \longrightarrow y$ be a classifier function for cyberbullying roles, where an input $p$ is assigned to $y$. Since $p$ is a pair of $c$ and $t$ that are individually assigned to $y^c$ and $y^t$, we present these pairs to a model with a single label that corresponds to the first text comment in the pair $p$ (\eg $f(\{c, t\}) \longrightarrow y^c$ and $f(\{t, c\}) \longrightarrow y^t$).
This reversal of $c$ and $t$ doubles the size of the corpus.
This is done to classify either $c$ or $t$ while presenting the other comment as a context.

\BfPara{Machine Learning Model} 
We use LLM model $\mathcal{E}$ to generate embeddings of any sample $x$, \ie
 $\mathcal{E}(\{w_1^{x}, w_2^{x}, w_3^{x}, \dots, w_l^{x}\}) \rightarrow \{{e}_1^{x}, e_2^{x}, e_3^{x}, \dots, e_l^{x}\}$.
Using the embeddings of the first token $\textsf{inp}={e}_1^{x}$ (\ie corresponding to the token $\langle\textsf{bos}\rangle$) or the average of all embeddings ($\textsf{inp} = \frac{1}{l}\sum_i(e_i^x)$) as input to an ML model $f$, the model learns the cyberbullying roles of the input $x$.
Since we use $t$ and $c$ pairs, we obtain $\textsf{inp}$ as $\textsf{inp} = \textsf{inp}^{t} + \lambda~ \textsf{inp}^{c}$, where  $\lambda$ is a scaling constant for the effect of $c$ on the estimation of $y_x^{t}=f(\textsf{inp})$.
In our experiments, we use $\lambda=0.5$ to
train Random Forest, AdaBoost, and XGBoost models.

\BfPara{LLMs}
LLMs are also used to build role classification models.
For an input $\{t_i, c_i\}$, we present tokens
$\{w_1^{t_i}, w_2^{t_i}, \dots, w_l^{t_i}, \langle\textsf{sep}\rangle, w_1^{c_i}, w_2^{c_i}, \dots, w_l^{c_i}\}$
to learn $y_i^t$.
We modify the classification head of the LLM to scale the effect of the context $c_i$ on the output.
Considering $f(.)$ as consecutive $m$ blocks, \ie $f = b_m(b_{m-1}, \dots (b_0 (.)))$, where $b_m$ is the classification head block, we modify the input to $b_m$ to receive \\
$\{b_{m-1}(b_{m-2} \dots (b_0(\{w_1^{t_i}))) + \lambda~  b_{m-1}(b_{m-2} \dots (b_0(\{w_1^{c_i}))) \}$, where $\lambda=0.5$. We apply the same approach when fine-tuning the models. 

\begin{table}[t]
\fontsize{7}{9}\selectfont
\centering
\resizebox{1\linewidth}{!}{%
    \begin{tabular}{p{8cm}cc}
    \toprule
    \textbf{Comment Text} & \textbf{Role} & \textbf{Harm} \\ 
    \midrule
    \textbf{Q:} [Well your the true b\censor{itch}]\textsuperscript{Gen. Insult} no it not a joke when someone cuts but why would you cut in the first place [b\censor{itch}]\textsuperscript{Gen. Insult} & Harasser & 1 \\
    \textbf{A:} [Are you really doing this right now?]\textsuperscript{Assertive Self. Def.} [Please leave me alone.]\textsuperscript{Assertive Self. Def.} & Victim & 1 \\
    \hline
    
    \textbf{Q:} [YOURE NOT EVEN UGLY OMFGGGGGGGGGGGGGGGGG]\textsuperscript{Good Characteristics} [I SHALL STAB THAT PERSON]\textsuperscript{Threat/Blackmail} OXOXOXOXOXOOXO & {\centering \makecell{Bystander\\Defender}} &  2 \\
    \textbf{A:} ILY XOOXXOXOXO & Bystander Other & 0 \\
    \bottomrule
    \end{tabular}
    }
\caption{Samples of Q\&A pairs and their labels.}
\label{tab:dataset_example}
\end{table}

\begin{table}[t]
\centering
\scalebox{0.9}{%}{\linewidth}{!}{
\begin{tabular}{llcccc}
\toprule
\multirow{2}{*}{Class} & \multirow{2}{*}{\# Comments (\%)} & \multicolumn{4}{c}{Median Length (Tokens)} \\ \cmidrule{3-6} 
                    &                 & RoBERTa & GPT-2 & T5 & BERT \\ \midrule
\textbf{0}            & 3574   (\textbf{2.918}\%)  & 11      & 9     & 13 & 12   \\
1              & 1354   (\textbf{1.105}\%)  & 15      & 13    & 17 & 16   \\
\textbf{2}  & 424    (\textbf{0.346}\%)  & 32      & 30    & 38 & 33   \\
\textbf{3} & 24     (\textbf{0.020}\%)  & 17.5    & 15.5  & 20 & 16.5 \\
\textbf{4}    & 117126  (\textbf{95.612}\%) & 9       & 7     & 10 & 10   \\ \midrule
\textbf{Total}               & 122502 (\textbf{100\%})    & \textbf{9}       & \textbf{7}     & \textbf{10} & \textbf{10}  \\
\bottomrule
\end{tabular}}
\caption{AMiCA dataset statistics and comments lengths. 
Classes $0$, $1$, $2$, $3$, and $4$ representing \textit{harasser}, \textit{victim}, \textit{bystander-defender}, \textit{-assistant}, and \textit{-other}, respectively.}
\label{tab:dataset_statistics}
\end{table}

\begin{figure}[t]
\centering
  \includegraphics[width=0.55\linewidth]{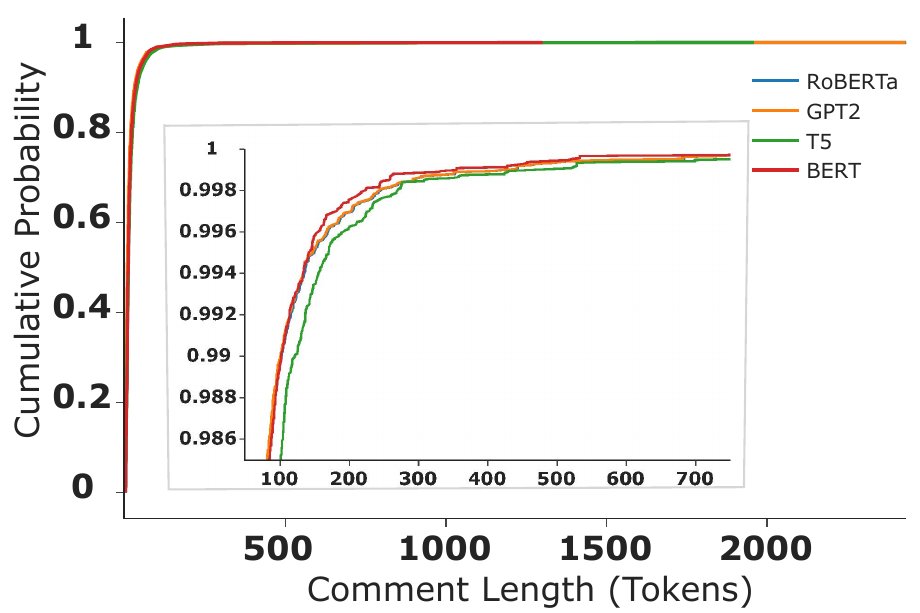}
  \caption{ECDF of comment lengths using various tokenizers with zoomed-in chart in the center. The max-length is set as the $99${-th} percentile, \ie 103, 101, 121, and 101 tokens for
  RoBERTa,
GPT2,
T5, and
BERT tokenizers, respectively.}
  \label{fig:comment_lengths}
\end{figure}

\begin{figure}[!ht]
\centering
  \includegraphics[width=0.90\linewidth]{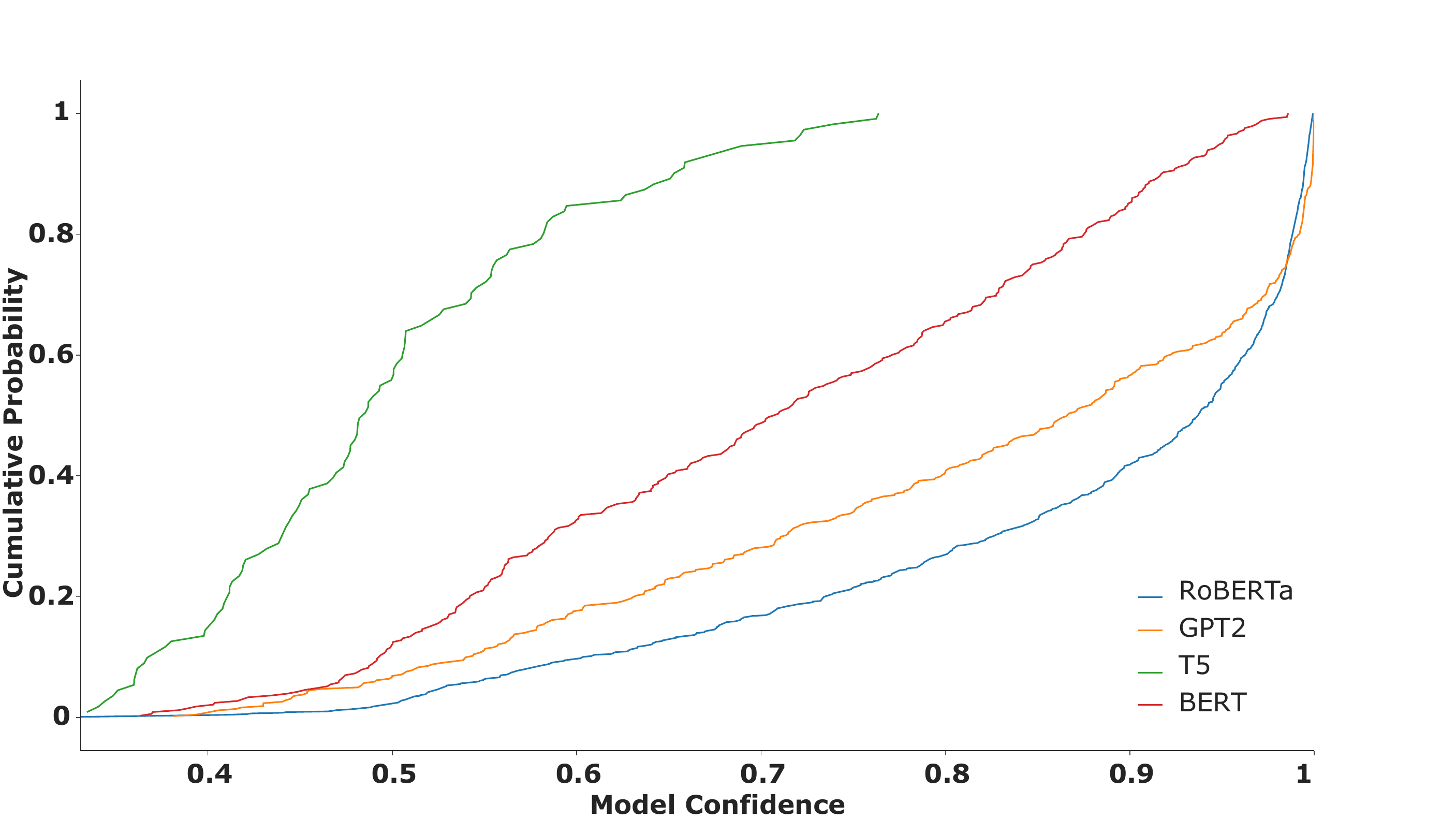}
  \caption{ECDF of model confidence for correctly identified victim comments for fine-tuned LLMs on oversampled data. $25${-th} percentile thresholds are:
  $0.7863$, $0.6737$, $0.4198$, and $0.5611$ for RoBERTa, GPT2, T5, and BERT, respectively.}
  \label{fig:correct_preds_cdf}
\end{figure}

\BfPara{Data Collection and Processing}
The AMiCA dataset \cite{van2018automatic} is, to our knowledge, the only available dataset of adequate size that is explicitly labeled with cyberbullying roles. 
This dataset was gathered from ASKfm, a social networking site where users can anonymously ask and answer questions. The Q\&A pairs were collected by crawling through a set of seed profiles between April and October 2013. The corpora contains $113,698$ English posts and $78,387$ Dutch posts, of which we excluded the Dutch posts. The dataset includes role labels for each question and answer, and sub-sentence behavior labels. 
An example of these annotations is provided in Table~\ref{tab:dataset_example}. Table~\ref{tab:dataset_statistics} provides an overview of the distribution of classes in the dataset.
Although the dataset does contain detailed annotations including the cyberbullying severity and labeled text-spans within each comment, we chose not to utilize this additional information in training our models and instead rely solely on the raw text of the comments, given that this information would not be readily available in real-world scenarios.
The anonymity and nature of AMiCA make it impossible to utilize a large amount of context when identifying roles, and the small dataset size makes it difficult to train large models effectively. We investigate several strategies to overcome this challenge. \\

\BfPara{Transforming Q\&A to Context-Target}
The AMiCA dataset is structured as Q\&A pairs with labels for each comment. 
To maintain the connection between the comments, we generate two samples for each Q\&A pair in the original dataset, utilizing both labels. In the first sample, the question is used as the context and the answer as the target for role identification. In the second sample, the answer is used as the context and the question as the target.
Therefore, the original $61,251$ English Q\&A pairs are transformed into $122,502$ samples. 
Because $95$\% of comments contain \textit{bystander-other} responses, we randomly sample $5,000$ pairs with %DLH note: the?
\textit{bystander-other} label.

\BfPara{Data Imbalance\space in Latent Space}%{Oversampling Embeddings}
We use ADASYN~\cite{adasyn} with its default parameters and $n\text{-neighbors}=15$ to oversample the minority classes, \textit{Harasser}, \textit{Victim}, \textit{Bystander Defender}, and \textit{Bystander Assistant}, in the latent space.

Providing raw samples to LLMs to obtain representations for ADASYN,
we use the following models, BERT~\cite{devlin2019bert}, RoBERTa~\cite{loureiro2022TwitterRoBERTa} base model (\ie trained and fine-tuned on $124$ million tweets for sentiment analysis), T5~\cite{raffel2023T5}, and GPT-2~\cite{RadfordGPT2}. 
Using the respective tokenizer, LLMs tokenize each comment separately to a specific length that is determined by the $99${-th} percentile of the length of the observed comment in the data. 
Comment length varies widely, but the vast majority of comments are short with the median comment length being only $10$ tokens when using RoBERTa's tokenizer (see Table~\ref{tab:dataset_statistics}). \textit{Bystander-defender} comments tend to be the longest, with a median length of 32 tokens, while the \textit{bystander-other} comments, which are the least related to cyberbullying instances, are the shortest, with a median length of only $9$ tokens. 
The empirical cumulative distribution function (ECDF) of the length of the comment is shown in Figure~\ref{fig:comment_lengths}.

Comment pairs are padded or truncated to the same length, and then forwarded together to the model to obtain the final embeddings.
For example, 
an input pair $\{t_i, c_i\}$ with the $99${-th} percentile of length $l$ is tokenized and passed as \\
$\{w_1^{t_i}, w_2^{t_i}, \dots, w_l^{t_i}, \langle\textsf{sep}\rangle, w_1^{c_i}, w_2^{c_i}, \dots, w_l^{c_i}\}$, where $w_1^{t_i}$ and $w_1^{c_i}$ are the $\langle\textsf{bos}\rangle$ tokens for $t_i$ and $c_i$, respectively.
The final representation is the weighted sum vector of the $t_i \text{ and } c_i$ embeddings, which is calculated as $e_1^{t_i} + \lambda~ e_1^{c_i}$, where $e_1^{t_i}$ and $e_1^{c_i}$ correspond to $w_1^{t_i}$ and $w_1^{c_i}$, respectively, and $\lambda$ is a factor to balance their weights ($\lambda$=$0.5$ in our experiments). 

\BfPara{Model Selection}
For role detection in cyberbullying interactions, we consider six models to train on two different types of data obtained using four different types of embedding strategies. 
Using four LLMs, \ie BERT~\cite{devlin2019bert}, RoBERTa~\cite{loureiro2022TwitterRoBERTa}, T5~\cite{raffel2023T5}, and GPT-2~\cite{RadfordGPT2}, we train various ML models (\ie Random Forest, AdaBoost, and XGBoost), and fine-tune and ensemble LLMs to perform the role detection task. Next, we outline the settings and the fine-tuning process.

\ul{\textbf{Machine Learning Models:}} Random Forest is constructed with 300 decision trees grown to the maximum extent and without bootstrapping (\ie all samples used for each tree). The final output is done by a majority vote.
The AdaBoost classifier starts by training a decision tree (with a maximum depth of $1$) using the dataset and then trains additional trees after adjusting the weights of incorrectly classified samples, so more attention is given to difficult cases.
The XGBoost classifier uses $100$ boosted decision trees grown to a depth of 6 and trained using a uniform sampling method on half the dataset for each boosting iteration with a learning rate of $0.3$. The tree construction algorithm is the faster histogram optimized approximate algorithm with $256$ histogram bins.

\begin{table*}[!ht]
\centering
\scalebox{0.9}{
\begin{tabular}{llccccc|ccccc} 
\toprule
\multicolumn{1}{l}{\multirow{2}{*}{LLM}} & \multirow{2}{*}{Model}        & \multicolumn{5}{c|}{Overall Metrics}                                                              & \multicolumn{5}{c}{Metrics with Threshold}                                                     \\ 
\cmidrule{3-12}
\multicolumn{1}{c}{}                     &                               & A                  & R                & P                  & F1                & Top-2 F1          & A                 & R               & P                  & F1                & RR             \\ 
\midrule
\multirow{3}{*}{Baseline}                & Base-RoBERTa \cite{jacobs2022automatic}                  & 0.676          & 0.676             & 0.509             & 0.580              & 0.669                & ---              & ---              & ---                & ---               & ---                \\
                                         & OffensEval Filtered \cite{rathnayake-etal-2020-enhancing}   & 0.540          & 0.540             & 0.641             & 0.561              & 0.623                & ---              & ---              & ---                & ---               & ---                \\
                                         & OffensEval Unfiltered \cite{rathnayake-etal-2020-enhancing} & 0.264          & 0.264             & 0.258             & 0.217              & 0.282                 & ---              & ---              & ---                & ---               & ---                 \\

\midrule
\multirow{6}{*}{BERT}                    & Random Forest                 & 0.544             & 0.544             & 0.522             & 0.527             & 0.807             & 0.573             & 0.573             & 0.554             & 0.556             & 0.101          \\
                                         & XGBoost                       & 0.614             & 0.614             & 0.588             & 0.593             & 0.833             & 0.640             & 0.640             & 0.622             & 0.621             & 0.085          \\
                                         & AdaBoost                      & 0.340             & 0.340             & 0.430             & 0.368             & 0.651             & 0.429             & 0.429             & 0.533             & 0.463             & 0.284          \\
                                         & Classification Head           & 0.486             & 0.486             & 0.491             & 0.413             & 0.790             & 0.547             & 0.547             & 0.590             & 0.509             & 0.189          \\
                                         & Ensemble  & 0.444             & 0.444             & 0.498             & 0.406             & 0.748             & 0.529             & 0.529             & 0.621             & 0.529             & 0.259                              \\
                                         & Fine-tuned LLM                & \textbf{0.736}    & \textbf{0.736}    & \textbf{0.727}    & \textbf{0.721}    & \textbf{0.903}   & \textbf{0.761}    & \textbf{0.761}    & \textbf{0.757}    & \textbf{0.748}    & \textbf{0.069} \\ 
\midrule
\multirow{6}{*}{RoBERTa}                 & Random Forest                 & 0.507             & 0.507             & 0.502             & 0.496             & 0.785             & 0.534             & 0.534             & 0.528             & 0.523             & \textbf{0.112}  \\
                                         & XGBoost                       & 0.558             & 0.558             & 0.545             & 0.549             & 0.809             & 0.597             & 0.597             & 0.585             & 0.588             & 0.129           \\
                                         & AdaBoost                      & 0.388             & 0.388             & 0.429             & 0.404             & 0.706             & 0.436             & 0.436             & 0.478             & 0.452             & 0.175           \\
                                         & Classification Head           & 0.662             & 0.662             & 0.700             & 0.664             & {0.910}           & 0.768             & 0.768             & {0.821}           & 0.775             & 0.303           \\
                                         & Ensemble  & 0.647             & 0.647             & 0.705             & 0.649             & 0.900             & 0.750             & 0.750             & 0.819             & 0.758             & 0.300                               \\
                                         & Fine-tuned LLM                & \textbf{0.835}    & \textbf{0.835}    & \textbf{0.834}    & \textbf{0.835}    & \textbf{0.957}    & \textbf{0.894}    & \textbf{0.894}    & \textbf{0.892}    & \textbf{0.893}    & 0.164           \\ 
\midrule 
\multirow{6}{*}{T5}                      & Random Forest                 & 0.521             & 0.521             & 0.516             & 0.512             & 0.784             & 0.550             & 0.550             & 0.545             & 0.540             & 0.109           \\
                                         & XGBoost                       & 0.581             & 0.581             & 0.570             & 0.574             & 0.811             & 0.621             & 0.621             & 0.610             & 0.614             & 0.133           \\
                                         & AdaBoost                      & 0.365             & 0.365             & 0.428             & 0.386             & 0.695             & 0.396             & 0.396             & 0.458             & 0.417             & 0.106           \\
                                         & Classification Head           & 0.541             & 0.541             & 0.559             & 0.527             & 0.834             & 0.628             & 0.628             & 0.707             & 0.626             & 0.278           \\
                                         & Ensemble  & 0.538             & 0.538             & 0.580             & 0.537             & 0.831             & 0.644             & 0.644             & \textbf{0.739}    & 0.654             & 0.322                               \\
                                         & Fine-tuned LLM                & \textbf{0.709}    & \textbf{0.709}    & \textbf{0.681}    & \textbf{0.669}   & \textbf{0.888}    & \textbf{0.718}    & \textbf{0.718}    & 0.700             & \textbf{0.681}    & \textbf{0.030}  \\ 
\midrule
\multirow{6}{*}{GPT2}                    & Random Forest                 & 0.512             & 0.512             & 0.495             & 0.496             & 0.794             & 0.546             & 0.546             & 0.532             & 0.530             & \textbf{0.114}  \\
                                         & XGBoost                       & 0.568             & 0.568             & 0.549             & 0.556             & 0.817             & 0.605             & 0.605             & 0.588             & 0.593             & 0.116           \\
                                         & AdaBoost                      & 0.340             & 0.340             & 0.430             & 0.368             & 0.651             & 0.429             & 0.429             & 0.533             & 0.463             & 0.284           \\
                                         & Classification Head           & 0.427             & 0.427             & 0.430             & 0.379             & 0.701             & 0.497             & 0.497             & 0.536             & 0.475             & 0.333           \\
                                         & Ensemble                      & 0.352             & 0.352             & 0.461             & 0.324             & 0.628             & 0.407             & 0.407             & 0.541             & 0.380             & 0.247           \\
                                         & Fine-tuned LLM                & {\textbf{0.737}} & \textbf{0.737}    & \textbf{0.739}    & \textbf{0.733}    & \textbf{0.896}    & \textbf{0.783}    & \textbf{0.783}     & \textbf{0.785}   & \textbf{0.780}     & 0.130            \\

\bottomrule
\end{tabular}}
\caption{Model performance of baselines and proposed models. The proposed models use oversampled data via ADASYN. \textbf{A}ccuracy, \textbf{R}ecall, \textbf{P}recision, \textbf{F1} score, and Rejection Rate (RR) are provided with/without confidence threshold using 10-fold stratified cross-validation. Best results per LLM are in bold.}
\label{tab:results}
\end{table*}

\ul{\textbf{Dense Neural Network Classification Head:}} A classification head is added on top of the LLM, which consists of a two-layer dense feedforward neural network. The first and second layers of this network have 2,048 and 1,024 units, respectively, and use ReLU activation functions. The network is then connected to a final softmax output layer, which is responsible for estimating the probability of each role.
To address the issue of overfitting, we use L2 regularization with a strength coefficient of 0.09 and a dropout rate of 0.3 for each layer.
The training process terminates when the validation loss ceases to decrease for 40 consecutive epochs.

\ul{\textbf{Ensemble of LLMs with Dense Classification Head:}} 
We build the ensemble classifier using the weights of dense classification heads trained with different epochs. 
The minimum number of training epochs to consider model weights in the ensemble is set to 275.
Then, we consider the weights after five-training-epoch intervals. The stopping condition is met once the validation loss stops decreasing for 40 epochs. The number of models are 26, 563, 25, and 776 (using augmented data), and 9, 22, 124, and 50 (using oversampled data) for BERT, RoBERTa, T5, and GPT-2, respectively.

\ul{\textbf{Fine-tuned LLM:}}
We fine-tune the last two blocks and the classification head of LLMs using our dataset to perform the role detection task.
The classification head consists of a single dense layer with 2048 units, a ReLU activation function, and L2 regularization with a strength coefficient of 0.09 and dropout with a rate of 0.3.
This layer is connected to the output layer with five units and a softmax activation function to generate the probabilities of each role.
The classification head is modified to receive the weighted sum of the $\mathsf{BOS}$ tokens of the target and context pair.

\ul{\textbf{Compute/GPU Settings:}} ML models are trained on a local workstation with an Intel Xeon Gold 6230R 2.1GHz CPU and 256 GB of RAM. The training and fine-tuning LLMs are conducted using GPU-enabled VMs on Colab and Lambda Cloud. The resources are selected for performance and efficiency and should not affect the results of the experiment.

\BfPara{Performance Evaluation}
We use weighted average F1-score and F1-scores per class to determine the performance of the model. We also record the overall accuracy and the weighted average of the recall and precision for each experiment. 
The confusion matrices of the models are analyzed to evaluate their performance and gain a better understanding of the challenges differentiating among distinct classes.
A common instance of confusion is mistaking harassers for victims and vice versa, but there are several other instances of overlap, such as harassers and bystander assistants, which have very similar behavior in cyberbullying interactions. 
This can be qualitatively explained as victims aggressively defending themselves, which can make it appear, especially with limited context, that they are the harassers. 
To provide a meaningful quantitative metric that accounts for these types of confusion, we also calculate thresholds for predicting a class and top-2 metrics. 

\BfPara{Prediction Threshold} Prediction threshold is adopted after the experiments are complete and the threshold used is standardized as the $25${-th} percentile of correctly classified victim comments. 
This cut-off is based on the ECDF of model confidence for correctly identified comments, shown in Figure \ref{fig:correct_preds_cdf}, and the fact that predicted victim comments typically have the lowest associated model confidence. 
This ensures thresholding is only applied when the model is relatively uncertain by its standards to predict the correct class. 
For thresholded metrics, an adjusted model prediction is used. If the probability of the model's original prediction exceeds the threshold or is correct, then the original prediction is used. If the model is not confident in its prediction and the probability is below the threshold, then the model's second choice class with the next highest probability is used instead. 
This provides a view of how often the correct class is in the top two choices of the model and shows how frequently two classes are confused with one another.

\begin{table*}[hbt!]
\centering
\scalebox{0.9}{
\begin{tabular}{llcccc|ccccc} 
\toprule
\multirow{2}{*}{Model}               & \multicolumn{1}{l}{\multirow{2}{*}{Class}} & \multicolumn{4}{c!{\vrule width \lightrulewidth}}{Overall Metrics} & \multicolumn{5}{c}{Metrics with Threshold}  \\ 
\cmidrule{3-11}
                                     & \multicolumn{1}{c}{}                       & R     & P     & F1    & Top-2 F1                                   & R     & P     & F1    & RR    & Support     \\

\midrule
\multirow{5}{*}{Random Forest}       & Harasser                                   & 0.479 & 0.507 & 0.493 & 0.850                                      & 0.521 & 0.559 & 0.539 & 0.136 & 3574        \\
                                     & Victim                                     & 0.195 & 0.322 & 0.243 & 0.462                                      & 0.242 & 0.377 & 0.295 & 0.175 & 1354        \\
                                     & Bystander Defender                         & 0.380 & 0.222 & 0.280 & 0.515                                      & 0.436 & 0.272 & 0.335 & 0.222 & 424         \\
                                     & Bystander Assistant                        & 0     & 0     & 0     & 0                                          & 0     & 0     & 0     & 0.375 & 24          \\
                                     & Bystander Other                            & 0.706 & 0.649 & 0.676 & 0.899                                      & 0.726 & 0.661 & 0.692 & 0.090 & 5000        \\ 
\midrule
\multirow{5}{*}{XGBoost}             & Harasser                                   & 0.599 & 0.575 & 0.587 & 0.886                                      & 0.655 & 0.631 & 0.643 & 0.169 & 3574        \\
                                     & Victim                                     & 0.312 & 0.399 & 0.350 & 0.611                                      & 0.390 & 0.495 & 0.436 & 0.243 & 1354        \\
                                     & Bystander Defender                         & 0.410 & 0.290 & 0.340 & 0.557                                      & 0.467 & 0.357 & 0.404 & 0.231 & 424         \\
                                     & Bystander Assistant                        & 0     & 0     & 0     & 0                                          & 0     & 0     & 0     & 0.208 & 24          \\
                                     & Bystander Other                            & 0.735 & 0.736 & 0.736 & 0.912                                      & 0.768 & 0.762 & 0.765 & 0.105 & 5000        \\ 
\midrule
\multirow{5}{*}{AdaBoost}            & Harasser                                   & 0.279 & 0.416 & 0.334 & 0.728                                      & 0.343 & 0.500 & 0.407 & 0.167 & 3574        \\
                                     & Victim                                     & 0.182 & 0.162 & 0.171 & 0.401                                      & 0.219 & 0.211 & 0.215 & 0.168 & 1354        \\
                                     & Bystander Defender                         & 0.474 & 0.127 & 0.200 & 0.494                                      & 0.479 & 0.131 & 0.206 & 0.127 & 424         \\
                                     & Bystander Assistant                        & 0     & 0     & 0     & 0                                          & 0     & 0     & 0     & 0.125 & 24          \\
                                     & Bystander Other                            & 0.567 & 0.582 & 0.574 & 0.836                                      & 0.619 & 0.624 & 0.621 & 0.172 & 5000        \\ 
\midrule
\multirow{5}{*}{Classification Head} & Harasser                                   & 0.755 & 0.654 & 0.701 & 0.965                                      & 0.828 & 0.739 & 0.781 & 0.204 & 3574        \\
                                     & Victim                                     & 0.376 & 0.568 & 0.452 & 0.805                                      & 0.495 & 0.690 & 0.576 & 0.278 & 1354        \\
                                     & Bystander Defender                         & 0.509 & 0.579 & 0.542 & 0.744                                      & 0.620 & 0.709 & 0.661 & 0.323 & 424         \\
                                     & Bystander Assistant                        & 0     & 0     & 0     & 0.074                                      & 0.042 & 0.100 & 0.059 & 0.292 & 24          \\
                                     & Bystander Other                            & 0.878 & 0.883 & 0.881 & 0.955                                      & 0.904 & 0.902 & 0.903 & 0.084 & 5000        \\ 
\midrule
\multirow{5}{*}{Ensemble}            & Harasser                                   & 0.728 & 0.657 & 0.691 & 0.968                                      & 0.797 & 0.742 & 0.769 & 0.198 & 3574        \\
                                     & Victim                                     & 0.393 & 0.562 & 0.462 & 0.803                                      & 0.488 & 0.656 & 0.560 & 0.261 & 1354        \\
                                     & Bystander Defender                         & 0.588 & 0.560 & 0.574 & 0.743                                      & 0.654 & 0.642 & 0.648 & 0.294 & 424         \\
                                     & Bystander Assistant                        & 0.045 & 0.048 & 0.047 & 0.077                                      & 0.045 & 0.063 & 0.053 & 0.364 & 24          \\
                                     & Bystander Other                            & 0.882 & 0.881 & 0.881 & 0.955                                      & 0.909 & 0.894 & 0.901 & 0.074 & 5000        \\ 
\midrule
\multirow{5}{*}{Fine-tuned LLM}      & Harasser                                   & 0.824 & 0.802 & 0.813 & 0.967                                      & 0.894 & 0.875 & 0.885 & 0.204 & 3574        \\
                                     & Victim                                     & 0.654 & 0.665 & 0.660 & 0.907                                      & 0.775 & 0.805 & 0.790 & 0.310 & 1354        \\
                                     & Bystander Defender                         & 0.639 & 0.727 & 0.680 & 0.831                                      & 0.759 & 0.826 & 0.791 & 0.344 & 424         \\
                                     & Bystander Assistant                        & 0     & 0     & 0     & 0.167                                      & 0.042 & 0.091 & 0.057 & 0.458 & 24          \\
                                     & Bystander Other                            & 0.913 & 0.916 & 0.914 & 0.978                                      & 0.941 & 0.937 & 0.939 & 0.079 & 5000        \\
\bottomrule
\end{tabular}}
\caption{Per-class results for best models of each type. The base LLMs used are BERT for Random Forest and XGBoost, GPT-2 for AdaBoost, and RoBERTa for Classification Head, Ensemble, and Fine-tuned LLM. Augmented data generally yields the best results, although, in the case of RoBERTa Fine-tuned LLM, the best results are achieved using oversampled data.}
\label{tab:per-class_results}
\end{table*}

\begin{table*}[hbt!]
\centering
\scalebox{0.85}{
\begin{tabular}{m{5.2cm}m{5.2cm}cccc} 
\toprule
Question                                                                                                                                & Answer                                                                                                                                                                                                                               & \begin{tabular}[c]{@{}c@{}}Q \\ 
    Pred. \\ 
    Role\end{tabular} & \begin{tabular}[c]{@{}c@{}}Q \\ 
    Actual \\ 
    Role\end{tabular} & \begin{tabular}[c]{@{}c@{}}A \\ 
    Pred. \\ 
    Role\end{tabular} & \begin{tabular}[c]{@{}c@{}}A \\ 
    Actual\\ 
    Class\end{tabular}  \\ 
\midrule
come visit me \censor{fgt} & busy \censor{fgt} & 4 & 0 & 0 & 1 \\
\specialrule{0.2pt}{0.1\jot}{0.1pc} sit on my face & \censor{SLUT} & 1 & 0 & 0 & 0 \\
\specialrule{0.2pt}{0.1\jot}{0.1pc} wants some butter u \censor{FUCKING} CRACKER & Who the \censor{fuck} even puts butter on crackers you \censor{dickhole} & 0 & 0 & 0 & 1 \\
\specialrule{0.2pt}{0.1\jot}{0.1pc} Pretty sure you told me you weren't a virgin awks & Pretty sure you \censor{fucked} ur dad awks & 1 & 0  & 0 & 1 \\
\bottomrule
\end{tabular}}
\caption{Examples of harasser and victim comment confusion.}
\label{tab:comment_confusion}
\end{table*}

\section{Experiments and Results} \label{sec:results}
In this section, we compare the results of the proposed models and the models implemented to serve as baselines. All models were evaluated on the same AMiCa \cite{van2018automatic} dataset. 

\BfPara{Baseline 1: OffensEval}  OffensEval \cite{rathnayake-etal-2020-enhancing} is an LLM based model composed of three DistilBERT models functioning as an ensemble for cyberbullying role classification. In the original authors' configuration of the dataset, each question and answer is a standalone post/training sample. The function of the `outer' model is to determine whether or not a post is cyberbullying, \ie binary classification. If the post is classified as bullying, the `bully' model determines whether the role is Harasser or Bystander Assistant. Likewise, if the outer model determines a post is not bullying, 
the `defender' model determines if the role is Victim or Bystander Defender. To overcome class imbalance, this approach uses 10 stratified fold cross validation paired with weighted random sampling when training the outer model. To the best of our understanding, OffensEval was not trained to classify Bystander Other. To more comprehensively evaluate OffensEval with our approaches, which consider the full set of 5 roles in the dataset, we considered two approaches, OffensEval Filtered and OffensEval Unfiltered. For OffensEval Filtered we trained and tested the model on a reduced dataset with only the samples of roles Harasser, Victim, Bystander Defender, and Bystander Assistant. This approach directly replicates the model implemented by the original authors as they did not consider the Bystander Other role. For OffensEval Unfiltered, we expanded the testing set of each fold to include Bystander Other. This second approach enables a more direct comparison with the approaches we propose. Observe that considering real-world scenarios, OffensEval Filtered assumes that existence of a previous model that will filter the Bystander Other interactions from the dataset.

\BfPara{Baseline 2: Base-RoBERTa} To serve as a baseline for comparison, we implemented the RoBERTa-based model proposed in \cite{jacobs2022automatic}. The authors of this paper stated that this model achieved an F1 score of $0.6$. To the best of our understanding, the original authors interpreted the `Bystander Other' role as their `Not Bullying' class. Moreover, due to the significant class imbalance presented by the few occurrences of `Bystander Assistant', they merged `Bystander Assistant' into `Harasser'.

\BfPara{Model Performance}
Table~\ref{tab:results} shows the performance of different models trained using oversampled, as well as the performance for the baseline models. Regarding the performance of baseline methods, we can observe that SAC-LR achieved a F1 score of $67$\%, SAC-SVM obtained a F1 score of $65$\%, OffensEval filtered obtained an F1 score of $53$\%, and finally, OffensEval unfiltered, \ie OffensEval predicting all 5 classes, achieved an F1 score of $21$\%.
Of our proposed methods, fine-tuning the last two blocks of RoBERTa proved to be uniformly the best model for every metric when using oversampling. It achieves an accuracy of $83.5$\%, which increases roughly $6$\% after thresholding with a rejection rate of $16.4$\%. This model also outperforms all the baseline methods.
Generally, there is a notable increase in the top-2 F1 score compared to the F1 score. This indicates the challenge that the model encounters in accurately determining the top prediction and resolving confusion. Applying a confidence threshold helps the model reject uncertain predictions due to either limited context or class ambiguity.

\BfPara{Per-class Model Performance} Detailed per-class metrics are also reported in Table~\ref{tab:per-class_results} for the model that performs the best for each type, measured by the weighted average F1 score without thresholding. 
Although there is some variation among models, generally, the \textit{bystander-other} and \textit{harasser} comments are the easiest to distinguish, with similar levels of recall and precision and high F1 scores around $0.9$ and $0.8$ for the best models, respectively. 
\textit{Victim} and \textit{bystander defender} comments lie in the middle with F1 scores slightly below $0.7$. 
Although the sample size of \textit{victim} comments is larger, they prove to be slightly more difficult to recognize compared to the comments of \textit{bystander defender}. This could be due to the observed overlap between victims and the more frequent harasser class when victims defend themselves using aggressive language. 
Most models tend to ignore or incorrectly recognize the \textit{bystander-assistant} class. 
This is likely due to a combination of only having $24$ total occurrences of the role in the entire dataset and the similarity of this role to harassers, particularly without more available context. 
The top-2 F1 scores per-class show that the models frequently choose between two similar classes. Although the top-1 F1 scores for victims and bystander defenders are $0.66$ and $0.68$, respectively, they increase to $0.91$ and $0.83$ when considering the top-2 predicted classes. The increase in F1 scores is the largest by far for victims, most likely due to their confusion with harassers. The top-2 metrics provide insight into the potential performance of the proposed models when using different datasets that have more diverse/distinct classes or clearer guidelines for annotating samples, \eg when aggressive victims use the same language as harassers.

\section{Discussion}
\BfPara{Class Confusion}
The fine-tuned RoBERTa model performs well in most situations, but struggles in some surprising situations such as distinguishing between harassers and victims. For seemingly opposite roles, having $8.22$\% of harasser comments mistakenly labeled as victims and $23.12$\% of victims labeled as harassers is unexpected. 
Table~\ref{tab:comment_confusion} shows a sample of Q\&A pairs with harassers and victims mistaken for one another. 
In several cases, victims mimic the language used by the harasser in their defensive response or otherwise aggressively defend themselves with language typically associated with harassment. 
This makes it difficult, even for a human familiar with the subject reading these comments, to identify with a high degree of certainty who the victim is in these interactions.
Although the exact reasoning behind the model's struggles remains unclear, by analyzing the model's mistakes, it is clear that the task is not trivial and that before substantial improvement in role prediction performance can be achieved, we likely need to either clarify role definitions or expand the number of possible roles for each comment.

\BfPara{Model Application}
While the proposed models consider the Ask.fm Question-Answer pair format, the models could be easily applied to data formats used in other platforms. For instance, a majority of platforms utilize a thread based discussion focused around a single post, \eg Reddit, Youtube, Facebook, and Instagram. To adapt the thread-based format, we could build a Q-A pair using the initial post text as the question and each individual comment as an answer. Another approach would be to consider each instance of a user mention, often seen as one user tagging another via the `@' symbol. Each instance of a comment with a user mention and the response by the mentioned user can constitute a Q-A pair.

\section{Conclusion} \label{sec:conclusion}
Our work investigates cyberbullying role detection in social media interactions using an imbalanced dataset with five classes (AMiCA dataset), \ie  Harasser, Victim, Bystander Assistant, Bystander Defender, and Bystander Other. The issue of cyberbullying has frequently posed a challenge, with over half of adolescents reporting instances of bullying while using social networks or engaging in online chats. Having a model for identifying the roles of cyberbullying instances would be beneficial for adolescents and parents, as this model could enable the implementation of more effective anti-bullying tools. Previous studies have focused on determining whether a post exhibits bullying behavior or not. More recently, some research has begun exploring the detection of different roles involved in cyberbullying, including victim, bully, and bystander. Contributing to the detection of cyberbullying roles in social media comments, this study explores the performance of various models with different training strategies and sheds light on the strengths and shortcomings of the employed methods. We plan to publish the code upon paper acceptance.

An important task for future work is the development of a more comprehensive labeled dataset that enables better detection models by more accurately capturing the roles linked to a given comment. This could be achieved, for instance, by enabling the assignment of multiple roles to a single comment (\eg a single comment could be labeled as a Bystander Defender and a Harasser). Moreover, the labeling approach could be extended by including a degree of each role (\eg mild vs. severe harasser). Both mechanisms would help capture more accurately complex instances of role overlap.

\subsubsection{\ackname} This work was supported by NSF Awards \#2227488 and \#1719722 and a Google Award for Inclusion Research.

\balance
\bibliographystyle{splncs04}
\bibliography{ref.bib}

\begin{thebibliography}{10}
\providecommand{\url}[1]{\texttt{#1}}
\providecommand{\urlprefix}{URL }
\providecommand{\doi}[1]{https://doi.org/#1}

\bibitem{bastiaensens2015can}
Bastiaensens, S., Vandebosch, H., Poels, K., Van~Cleemput, K., DeSmet, A., De~Bourdeaudhuij, I.: ‘can i afford to help?’how affordances of communication modalities guide bystanders' helping intentions towards harassment on social network sites. Behaviour \& Information Technology  \textbf{34}(4),  425--435 (2015)

\bibitem{cheng_hierarchical_2019}
Cheng, L., Guo, R., Silva, Y., Hall, D., Liu, H.: Hierarchical attention networks for cyberbullying detection on the instagram social network. In: Proceedings of the 2019 {SIAM} International Conference on Data Mining ({SDM}), pp. 235--243. Society for Industrial and Applied Mathematics (2019). \doi{10.1137/1.9781611975673.27}

\bibitem{lu_cheng_2021_hant}
Cheng, L., Guo, R., Silva, Y.N., Hall, D., Liu, H.: Modeling temporal patterns of cyberbullying detection with hierarchical attention networks. ACM/IMS Trans. Data Sci.  \textbf{2}(2) (2021)

\bibitem{cheng_pi-bully_2019}
Cheng, L., Li, J., Silva, Y., Hall, D., Liu, H.: {PI}-bully: Personalized cyberbullying detection with peer influence. Electronic proceedings of {IJCAI} 2019 pp. 5829--5835 (2019)

\bibitem{MDavar2012}
Dadvar, M., {de Jong}, F., Ordelman, R., Trieschnigg, R.: Improved cyberbullying detection using gender information. In: Proceedings of the Twelfth Dutch-Belgian Information Retrieval Workshop (DIR 2012). pp. 23--25. Ghent University, Belgium (2012)

\bibitem{dadvar2018cyberbullying}
Dadvar, M., Eckert, K.: Cyberbullying detection in social networks using deep learning based models; a reproducibility study. arXiv preprint arXiv:1812.08046  (2018)

\bibitem{Dang_2022}
Dang, J., Liu, L.: Me and others around: The roles of personal and social norms in chinese adolescent bystanders’ responses toward cyberbullying. Journal of Interpersonal Violence  \textbf{37}(9-10),  NP6329--NP6354 (2022). \doi{10.1177/0886260520967128}, \url{https://doi.org/10.1177/0886260520967128}, pMID: 33073678

\bibitem{devlin2019bert}
Devlin, J., Chang, M.W., Lee, K., Toutanova, K.: Bert: Pre-training of deep bidirectional transformers for language understanding (2019)

\bibitem{Hamlett_Powell_Silva_Hall_2022}
Hamlett, M., Powell, G., Silva, Y.N., Hall, D.: A labeled dataset for investigating cyberbullying content patterns in instagram. Proceedings of the International AAAI Conference on Web and Social Media  \textbf{16}(1),  1251--1258 (May 2022). \doi{10.1609/icwsm.v16i1.19376}, \url{https://ojs.aaai.org/index.php/ICWSM/article/view/19376}

\bibitem{adasyn}
He, H., Bai, Y., Garcia, E.A., Li, S.: Adasyn: Adaptive synthetic sampling approach for imbalanced learning. In: 2008 IEEE International Joint Conference on Neural Networks (IEEE World Congress on Computational Intelligence). pp. 1322--1328 (2008). \doi{10.1109/IJCNN.2008.4633969}

\bibitem{jacobs2022automatic}
Jacobs, G., Van~Hee, C., Hoste, V.: Automatic classification of participant roles in cyberbullying: Can we detect victims, bullies, and bystanders in social media text? Natural Language Engineering  \textbf{28}(2),  141--166 (2022)

\bibitem{loureiro2022TwitterRoBERTa}
Loureiro, D., Barbieri, F., Neves, L., Anke, L.E., Camacho-Collados, J.: Timelms: Diachronic language models from twitter (2022)

\bibitem{Orue2023}
Orue, I., Fernández-González, L., Machimbarrena, J.M., González-Cabrera, J., Calvete, E.: Bidirectional relationships between cyberbystanders’ roles, cyberbullying perpetration, and justification of violence. Youth \& Society  \textbf{55}(4),  611--629 (2023). \doi{10.1177/0044118X211053356}, \url{https://doi.org/10.1177/0044118X211053356}

\bibitem{RadfordGPT2}
Radford, A., Wu, J., Child, R., Luan, D., Amodei, D., Sutskever, I.: Language models are unsupervised multitask learners (2019)

\bibitem{raffel2023T5}
Raffel, C., Shazeer, N., Roberts, A., Lee, K., Narang, S., Matena, M., Zhou, Y., Li, W., Liu, P.J.: Exploring the limits of transfer learning with a unified text-to-text transformer (2023)

\bibitem{rathnayake-etal-2020-enhancing}
Rathnayake, G., Atapattu, T., Herath, M., Zhang, G., Falkner, K.: Enhancing the identification of cyberbullying through participant roles. In: Proceedings of the Fourth Workshop on Online Abuse and Harms. pp. 89--94. Association for Computational Linguistics (2020)

\bibitem{salmivalli_participant_2014}
Salmivalli, C.: Participant roles in bullying: How can peer bystanders be utilized in interventions? Theory Into Practice  \textbf{53}(4),  286--292 (2014)

\bibitem{salmivalli_bullying_1996}
Salmivalli, C., Lagerspetz, K., {Björkqvist}, K., {Österman}, K., Kaukiainen, A.: Bullying as a group process: Participant roles and their relations to social status within the group. Aggressive Behavior  \textbf{22}(1),  1--15 (1996)

\bibitem{sanh2020distilbert}
Sanh, V., Debut, L., Chaumond, J., Wolf, T.: Distilbert, a distilled version of bert: smaller, faster, cheaper and lighter (2020)

\bibitem{singh_toward_2017}
Singh, V.K., Ghosh, S., Jose, C.: Toward multimodal cyberbullying detection. In: Proceedings of the 2017 {CHI} Conference Extended Abstracts on Human Factors in Computing Systems. pp. 2090--2099. Association for Computing Machinery (2017)

\bibitem{teng2023cyberbullying}
Teng, T.H., Varathan, K.D.: Cyberbullying detection in social networks: A comparison between machine learning and transfer learning approaches. IEEE Access  (2023)

\bibitem{van2018automatic}
Van~Hee, C., Jacobs, G., Emmery, C., Desmet, B., Lefever, E., Verhoeven, B., De~Pauw, G., Daelemans, W., Hoste, V.: Automatic detection of cyberbullying in social media text. PloS one  \textbf{13}(10),  e0203794 (2018)

\bibitem{wang_multi-modal_2020}
Wang, K., Xiong, Q., Wu, C., Gao, M., Yu, Y.: Multi-modal cyberbullying detection on social networks. In: 2020 International Joint Conference on Neural Networks ({IJCNN}). pp.~1--8 (2020)

\bibitem{xu_2012}
Xu, J.M., Jun, K.S., Zhu, X., Bellmore, A.: Learning from bullying traces in social media. In: Proceedings of the 2012 Conference of the North American Chapter of the Association for Computational Linguistics: Human Language Technologies. p. 656–666. NAACL HLT '12, Association for Computational Linguistics, USA (2012)

\bibitem{ziems_agg_rep_int_visb_2020}
Ziems, C., Vigfusson, Y., Morstatter, F.: Aggressive, repetitive, intentional, visible, and imbalanced: Refining representations for cyberbullying classification. CoRR  \textbf{abs/2004.01820} (2020), \url{https://arxiv.org/abs/2004.01820}

\end{thebibliography}
\end{document}